\documentclass{article}
\usepackage[utf8]{inputenc}
\usepackage{authblk}
\usepackage{xcolor}
\usepackage[english]{babel}
\usepackage{setspace}
\usepackage[margin=1.25in]{geometry}
\usepackage{graphicx}
\graphicspath{ {./figures/} }
\usepackage{subcaption}
\usepackage{amsmath}
\usepackage{lineno}
\usepackage{mathrsfs}
\usepackage{endnotes}
\usepackage{booktabs}
\usepackage{tabularx}
\usepackage{float}
\usepackage{CJKutf8}
\usepackage[colorlinks,linkcolor=blue,anchorcolor=blue,citecolor=blue]{hyperref}

\hypersetup{
    colorlinks=true,
    linkcolor=blue,
    urlcolor=blue 
}

\usepackage[style=nejm, 
citestyle=numeric-comp,
sorting=none]{biblatex}
\title{LadleNet: A Two-Stage UNet for Infrared Image to Visible  Image Translation Guided by Semantic Segmentation}

\author[1]{Tonghui Zou}
\author[2]{Lei Chen*}

\affil[1]{Xi'an Technological University, Xi'an, Shaanxi, 710021, China; zoutonghui@st.xatu.edu.cn}
\affil[2]{Xi'an Technological University, Xi'an, Shaanxi, 710021, China; clei@xatu.edu.cn}
\affil[*]{Corresponding Author.}

\date{}

\begin{document}

\begin{CJK}{UTF8}{gbsn}

\maketitle
\begin{abstract}
The translation of thermal infrared (TIR) images into visible light (VI) images plays a critical role in enhancing model performance and generalization capability, particularly in various fields such as registration and fusion of TIR and VI images. However, current research in this field faces challenges of insufficiently realistic image quality after translation and the difficulty of existing models in adapting to unseen scenarios. In order to develop a more generalizable image translation architecture, we conducted an analysis of existing translation architectures. By exploring the interpretability of intermediate modalities in existing translation architectures, we found that the intermediate modality in the image translation process for street scene images essentially performs semantic segmentation, distinguishing street images based on background and foreground patterns before assigning color information. Based on these principles, we propose an improved algorithm based on U-net called LadleNet. This network utilizes a two-stage U-net concatenation structure, consisting of Handle and Bowl modules. The Handle module is responsible for constructing an abstract semantic space, while the Bowl module decodes the semantic space to obtain the mapped VI image. Due to the characteristic of semantic segmentation, the Handle module has strong extensibility. Therefore, we also propose LadleNet+, which replaces the Handle module in LadleNet with a pre-trained DeepLabv3+ network, enabling the model to have a more powerful capability in constructing semantic space. The proposed methods were trained and tested on the KAIST dataset, followed by quantitative and qualitative analysis. Compared to existing methods, LadleNet and LadleNet+ achieved an average improvement of 12.4\% and 15.2\% in SSIM metrics, and 37.9\% and 50.6\% in MS-SSIM metrics, respectively. Particularly, LadleNet+ demonstrated state-of-the-art performance in terms of image clarity and perception. The source code will be made available at \url{https://github.com/Ach-1914/LadleNet/tree/main/}.
\end{abstract}

\section{Introduction}
In recent years, visual tasks based on thermal infrared (TIR) images have gained significant attention in the research domain. Examples of such tasks include semantic segmentation using TIR images \cite{in_1} and object detection using TIR images \cite{in_2}. In contrast to visible light imaging techniques, thermal imaging technology has the capability to capture thermal radiation that is imperceptible to the human eye, thereby providing additional information such as object textures and temperature radiation profiles. Furthermore, thermal imaging technology excels in capturing images under challenging conditions, such as darkness, rainy, foggy, and overexposed scenarios.However, relying solely on TIR images, which constitutes a unimodal information source, proves insufficient to achieve performance surpassing advanced visual tasks based on visible light (VI) images. TIR images also suffers from drawbacks like the absence of color and limitations in terms of human perception. Consequently, the prevailing direction in the field of computer vision involves the simultaneous utilization of information from both TIR and VI modalities for image enhancement. This approach aims to achieve performance exceeding that of advanced unimodal visual tasks. A multitude of studies have substantiated the feasibility of this approach \cite{in_3}.

The enhancement of Thermal Infrared (TIR) and Visible (VI) image primarily involves research directions focused on registration and fusion of TIR and VI images. The registration of TIR and VI images is analogous to conventional registration tasks, entailing the alignment of common objects across the two images to obtain registered versions. Nevertheless, unlike conventional registration tasks, TIR and VI images stem from distinct modalities, necessitating the alignment of heterogeneous images. This significantly escalates the complexity of the registration process. Presently, the most commonly employed registration algorithms in this domain are still traditional methods such as Scale-Invariant Feature Transform (SIFT) \cite{in_4} and Oriented FAST and Rotated BRIEF (ORB) \cite{in_5}. While these algorithms prove effective for general TIR and VI images registration, their efficacy often diminishes for registration tasks involving scenarios like nighttime, rainy, and overexposed conditions. Moreover, such traditional algorithms exhibit shortcomings in terms of poor generalization and unstable results, making real-time images registration challenging.

The fusion of TIR and VI images aims to generate images that combine the rich texture information from TIR images with the realistic color information from VI images. This not only enhances the performance of advanced visual tasks but also yields images that are more in line with human perception. However, when VI images lack sufficient color information, such as in nighttime or underexposed scenes, the resulting images quality becomes compromised. Additionally, due to the formidable challenge of creating paired datasets comprising images captured under the same scene conditions during nighttime, daytime, normal exposure, and underexposure scenarios in the real world, conventional model training methods struggle to provide accurate color information for images fusion in such situations. Furthermore, mainstream image fusion models currently rely on training with grayscale versions of color images \cite{in_6}. While TIR images are also grayscale, the disparate imaging principles often lead to unsatisfactory coloring outcomes.

Addressing the challenges present in the realm of Thermal Infrared (TIR) and Visible (VI) image enhancement, we posit that TIR-to-VI image translation stands as one of the most promising solutions. Fundamentally, TIR-to-VI image translation entails mapping TIR images to VI images through a network architecture. In this process, the model learns various mapping relationships between TIR and VI images, encompassing mappings in terms of color and texture. This approach augments image enhancement techniques by providing additional information, thereby enhancing model performance and generalization capabilities.

\begin{figure}[h]
    \centering
    \begin{subfigure}{0.9\textwidth}
        \includegraphics[width=0.9\textwidth, height=1.2in]{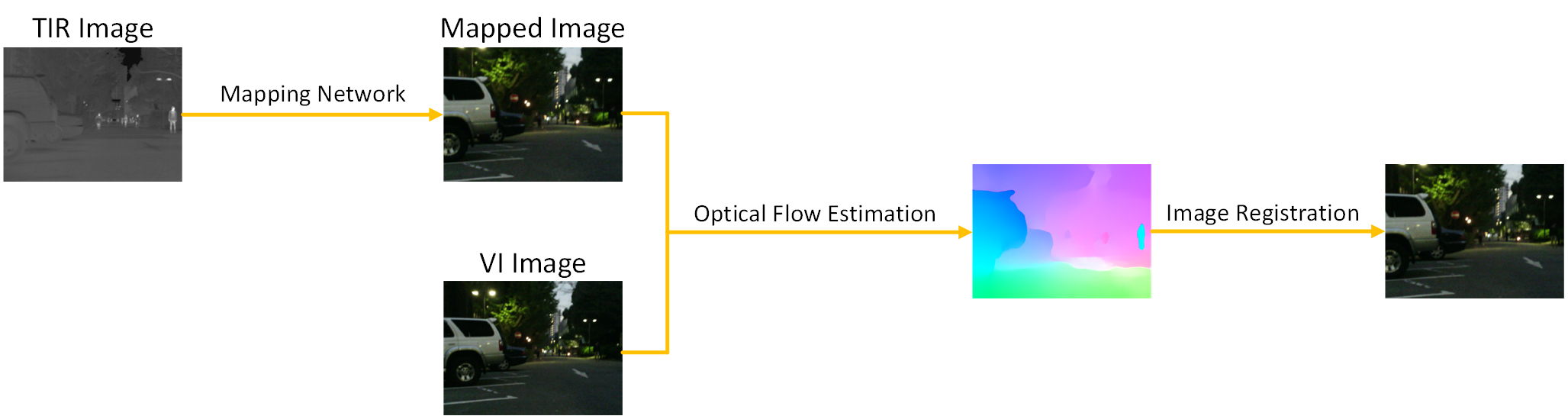}
        \caption{\label{fig:1a}}
    \end{subfigure}

    \vspace{0.5em}
    
    \begin{subfigure}{0.9\textwidth}
        \includegraphics[width=0.9\textwidth, height=1.2in]{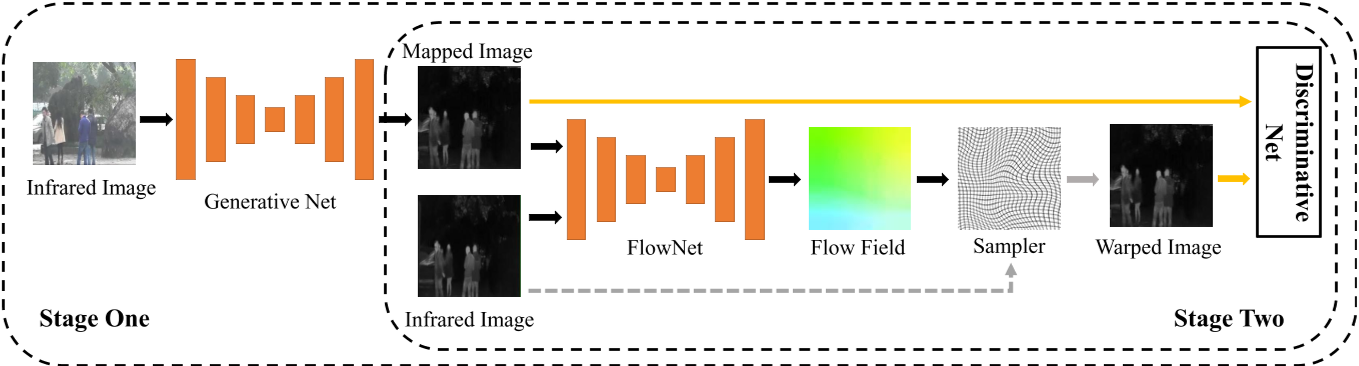}
        \caption{\label{fig:1b}}
    \end{subfigure}
    
    \caption{(\subref{fig:1a}) illustrates the process of image registration utilizing optical flow estimation, where the translated VI image obtained through mapping is aligned with the authentic non-aligned VI image. (\subref{fig:1b}) depicts the network proposed by Wang et al. for image registration based on optical flow estimation.}
    \label{fig:1}
\end{figure}

Specifically, within the context of TIR-to-VI image translation, a novel and efficient approach can be applied to aid TIR and VI image registration. This method involves translating non-aligned TIR images into corresponding VI images using the network, subsequently employing optical flow estimation methods \cite{in_7} to determine displacement disparities between the translated VI images and authentic VI images. This information guides the image registration process. The schematic representation of this approach is depicted in Figure \ref{fig:1a}. This methodology has been validated by Wang et al. \cite{in_8}, and their proposed method is illustrated in Figure \ref{fig:1b}. Additionally, this technique can be used to enhance the fusion of TIR and VI images when color information is insufficient. By providing information that approximates real colors, this method significantly improves fusion outcomes and performance in advanced visual tasks.

In fact, existing TIR-to-VI image translation methods still exhibit certain deficiencies in terms of image quality, rendering them unable to meet the high-precision mapping requirements for both texture and color, and lacking scalability. And the existing methods are not scalable, which makes it difficult to apply the model with flexibility in choosing the model according to the actual situation. In this paper, we propose a novel network architecture called LadleNet, which consists of a two-stage U-net concatenation. The initial U-net stage is referred to as the 'Handle' module, primarily focused on learning the mapping relationship between TIR and VI images while constructing an abstract semantic space. The secondary U-net stage is termed the 'Bowl' module, aiming to translate the abstract semantic space generated by the Handle module into realistically textured and colored images that closely resemble reality. This stage also aggregates shallow-level features from the Handle module, enhancing the fidelity of the generated images.

As the core function of the Handle module is to establish an abstract semantic space, we substitute the first-stage U-net with a pre-trained semantic segmentation network, thus extending LadleNet to create a more powerful model known as LadleNet+. The proposed LadleNet and LadleNet+ models are trained and evaluated on the KAIST dataset, achieving state-of-the-art performance compared to existing models. Additionally, we conduct ablation experiments that substantiate the effectiveness of our proposed approach. The primary contributions of this paper can be summarized as follows.

\begin{itemize}
    \item The proposed LadleNet architecture achieves superior performance compared to traditional methods for translating thermal infrared (TIR) images into visible light (VI) images. Compared to existing models, LadleNet demonstrates a significant advantage in image generation quality, with an average improvement of 12.4\% in SSIM and 37.9\% in MS-SSIM metrics.
    \item The scalable LadleNet+ architecture is introduced, innovatively employing a semantic-space-based approach for TIR to VI image translation. Compared to existing models, LadleNet+ exhibits an average improvement of 15.2\% in SSIM and 50.6\% in MS-SSIM metrics, achieving greater enhancement over LadleNet.
    \item To further validate the effectiveness of the semantic-space-based approach for TIR to VI image translation, we assess the quality of generated images by LadleNet and LadleNet+ architectures using additional metrics. LadleNet+ architecture consistently outperforms LadleNet and existing methods in terms of AG, MSE, VIF, and CC metrics, while LadleNet surpasses existing methods in AG, VIF, and CC metrics.
\end{itemize}

The remaining content of this paper is organized as follows. In Section 2, we present the relevant work pertaining to the scope of this study. Section 3 elaborates on the specific details of the proposed network architecture. In Section 4, the practical experiments conducted on the model are detailed. This section covers the datasets employed for model training and testing, specific implementation details of the network, comparative experiments among various models, and ablation experiments on the LadleNet model structure.Finally, Section 5 provides a comprehensive summary of the entire paper.

\section{Related work}
\subsection{U-net and Its Variations}
\begin{figure}[h]
    \centering
    \includegraphics[width=0.7\textwidth]{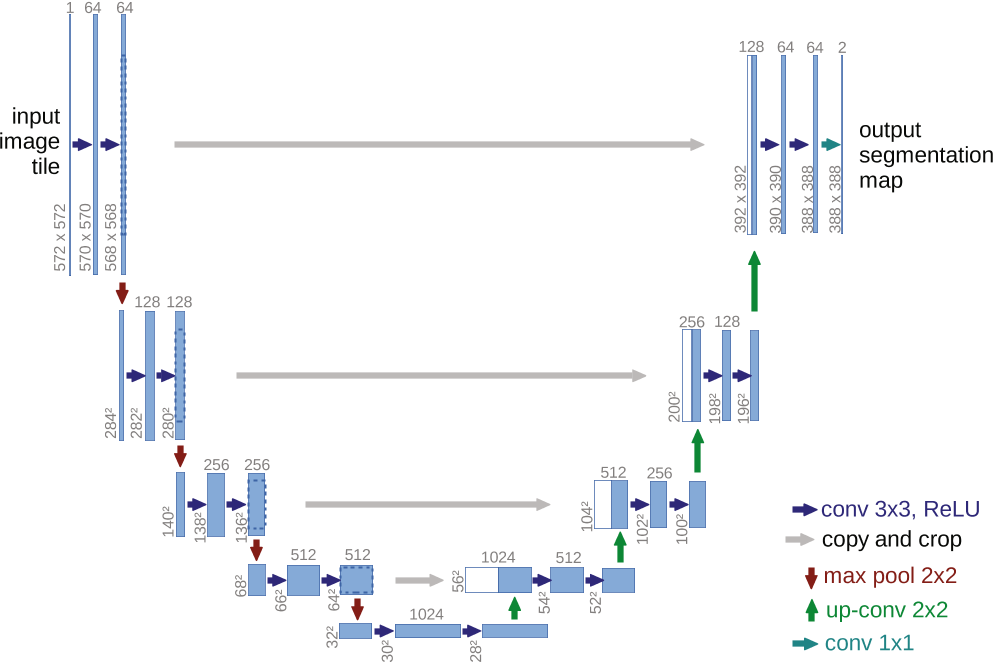}
    \caption{A basic architecture of U-net model.}
    \label{fig:2}
\end{figure}

The U-net network, introduced by Ronneberger et al. \cite{re_2}, is an image segmentation architecture. Building upon the traditional Auto-Encoder (AE) structure, the U-net network aggregates low-level features into high-level features, enabling the model to learn more intricate details. The structural diagram of the model is illustrated in Figure \ref{fig:2}. Initially employed for medical image segmentation, U-net's distinctive feature aggregation mechanism renders it particularly effective in segmenting small objects. Subsequently, the AE structure, often used in image generation and translation domains, was adopted in image translation tasks. In the later work by Isola et al. \cite{re_2}, the Pix2Pix network utilized the U-net as a generator for image translation. With increasing recognition of U-net's superiority, particularly in image translation, it gradually evolved into a benchmark model in this domain.

Due to its lightweight yet high-precision characteristics, researchers have increasingly explored advancements upon the U-net architecture, aiming to enhance its performance further. Zhang et al. \cite{re_3} introduced an improved U-net variant known as ResU-net, which leverages a structure similar to ResNet's residual blocks in place of the original U-net's convolutional layers. By incorporating skip connections, this architecture enhances the expressiveness and model performance of the U-net. Another approach to enhancement is the Bridged U-net proposed by Chen et al. \cite{re_5}. This network bridges two U-net structures through feature aggregation and incorporates additional skip connections, leading to a substantial performance boost.

\subsection{Image-To-Image Translation}
Image-to-image translation has emerged as a vibrant research domain in recent years. Within this field, various generative models are employed to translate images from one modality to another, resulting in increasingly realistic translated images. Notable models utilized in this domain include GANs \cite{re_6}, conditional GANs (cGANs) \cite{re_7}, and Pix2Pix \cite{re_2}.

TIR-to-VI image translation falls within the scope of image-to-image translation, yet it remains in its nascent stages. In 2018, Berg et al. \cite{re_8} presented a TIR-to-VI image translation network named TIR2Lab, based on an AE structure, at the IEEE/CVF Conference on Computer Vision and Pattern Recognition (CVPR). Liu et al. \cite{re_9} introduced a joint translation network IR2VI, incorporating a generator, a global discriminator, and an ROI discriminator. Although these models exhibited some performance shortcomings, they represented significant research contributions in this early stage of the field. Subsequently, Tao et al. \cite{re_10} employed a U-net framework for TIR-to-VI image translation, achieving notable improvements in model performance compared to previous studies. Bhat et al. \cite{re_11} proposed a TIR-to-VI image translation network based on the Pix2Pix architecture, utilizing a U-net generator and a cGAN framework, leading to enhanced model stability. However, these studies primarily focused on TIR image translation without leveraging additional information for enhanced translation. Yang et al. \cite{re_12} introduced a cGAN-based image translation method, GMA-CycleGAN, which capitalized on image semantic information. By converting TIR images to grayscale before mapping to VI images, this approach circumvented issues of edge blurriness.

Although advancements have been made in the TIR-to-VI image translation domain, various challenges persist. For instance, the simplicity of TIR2Lab and IR2VI models resulted in suboptimal translation outcomes. Translation methods based on U-net and Pix2Pix, while improving translation quality, often exhibit deficiencies in capturing image details. GMA-CycleGAN is a relatively effective approach, yet it suffers from color mapping errors and excessive model parameters, limiting its practical application and scalability. To address these challenges concurrently, we propose a more lightweight, efficient, and scalable network solution.

\section{Network Architectures}
Within the domain of Image-to-image translation, the most prominent research directions currently include models such as cGANs \cite{re_7} and Pix2Pix \cite{re_2}. These architectures exhibit exceptional performance in various image translation tasks. For TIR-to-VI image translation, however, such extensive and complex architectures may be redundant. On one hand, thermal imaging technology inherently lacks information about certain object features, making it highly unlikely for even powerful models to solely learn precise mapping relationships from image data. On the other hand, applications that necessitate TIR image translation, such as TIR and VI image registration and fusion, may not necessarily require exact mapping. Instead, a simple, fast, and efficient mapping network can better enhance outcomes following registration and fusion. Therefore, compared to complex GAN-based networks, a simpler and more efficient U-net based on the AE architecture is a better choice.

U-net is a network grounded in semantic information, initially applied in the field of medical image segmentation. Its distinct feature aggregation mechanism significantly enhances the model's capability to segment image details. For image translation tasks, U-net remains highly efficient, substantially improving performance compared to traditional AE architectures. In contrast to complex GAN networks, U-net is more stable and easier to converge, even achieving comparable performance in some domains. However, for intricate image translation tasks, U-net may encounter limitations. Consequently, researchers have explored improvements to U-net, with notable approaches including U-net++ \cite{net_1}, MDU-net \cite{net_2}, and ResU-net \cite{re_4}. Nevertheless, these enhancements primarily focus on image segmentation tasks and modifications to U-net's convolutional structure, rendering them non-scalable.

Among various U-net enhancement methods, Bridged U-net proposed by Chen et al. \cite{re_5} caught our attention. This network enhances segmentation performance by connecting two U-nets in series. Such an enhancement approach is more versatile, extending beyond segmentation tasks, and introduces a degree of scalability. Inspired by this method, we introduce the LadleNet model. This network modifies the original Bridged U-net's feature concatenation method and incorporates multiple skip connections to prevent model degradation. Furthermore, we propose the scalable LadleNet+ model, building upon the characteristic of constructing a semantic space in the first-stage U-net. The following sections present an introduction to the overarching concept and principles of the entire network architecture.

\subsection{LadleNet}
\begin{figure}[h]
    \centering
    \includegraphics[width=\textwidth]{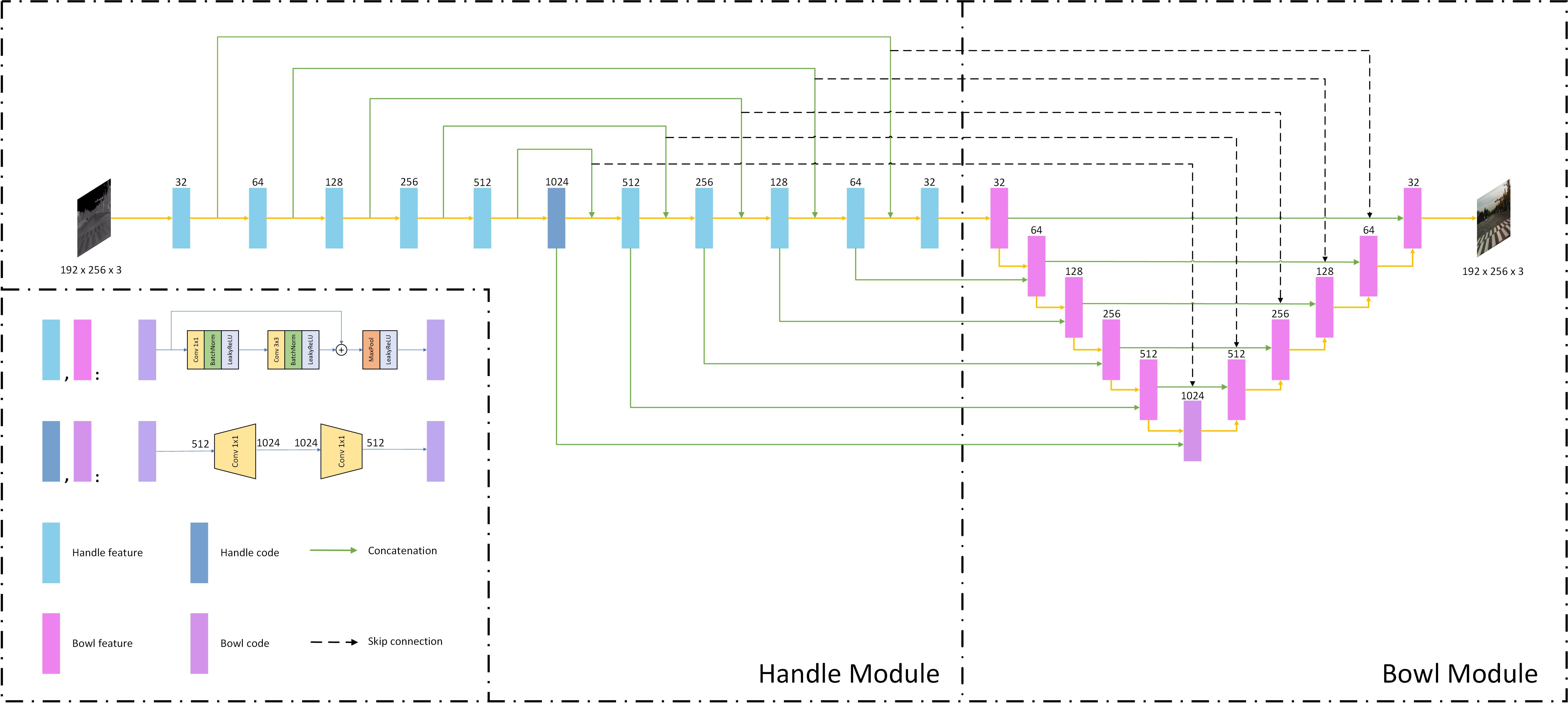}
    \caption{The structure of a basic LadleNet, which contains two parts, the Handle module (left side) and the Bowl module (right side).}
    \label{fig:3}
\end{figure}

LadleNet comprises two components: the "Handle" Module and the "Bowl" Module. The Handle module primarily conducts feature extraction, enhancing the model's expressive capacity by constructing a semantic space that maps TIR images to VI images. The Bowl module focuses on semantic space translation, achieving the translation of TIR images to VI images by extracting mapping relationships from the semantic space. Considering the model's complexity and parameter count, we initially constructed a foundational LadleNet based on the Bridged U-net architecture, as depicted in Figure \ref{fig:3}. The original Bridged U-net employs feature concatenation to link two U-nets, effectively averting the issue of model degradation caused by an excessive number of convolutional layers. However, this approach also somewhat diminishes the model's expressive capacity. To further enhance model performance, we modified the original connection method by directly chaining the two U-nets. Moreover, we introduced skip connections to each convolutional block, resulting in a substantial performance boost.

In addition to performance improvement, we also devised an extensible approach for LadleNet. Given that the Handle and Bowl modules serve distinct functions in the model, both possess strong extensibility. The following sections provide detailed introductions and analyses of these two modules.

\subsubsection*{Handle Module}
Within the foundational LadleNet architecture, the Handle module consists of a 5-layer U-net. Each convolutional block in the module comprises two 3x3 convolutional layers, responsible solely for feature extraction without downsizing feature maps. Within each convolutional block, a skip connection is established every two convolutional layers, followed by channel transformation using a 1x1 convolutional kernel, and subsequently, feature map dimension reduction via max-pooling. The "code" segment within the Handle module employs two 1x1 convolutional layers with 1024 channels. Its role is to construct a more abstract semantic space, thereby enhancing the effectiveness of features acquired by the decoder. In terms of feature aggregation, the original U-net structure is retained, with high-level features from the Handle module being conveyed to the shallow features of the Bowl module. This facilitates more efficient image translation within the Bowl module. Additionally, the Handle module introduces skip connections at the feature aggregation level, transmitting aggregated features from the Handle module to the aggregation layer of the Bowl module, thereby mitigating potential performance degradation arising from high model complexity.

\begin{figure}[h]
  \centering

  \begin{subfigure}{0.19\textwidth}
    \includegraphics[width=\linewidth]{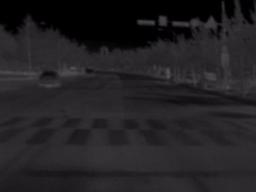}
    \caption{TIR Image}
    \label{fig:4a}
  \end{subfigure}
  \hfill
  \begin{subfigure}{0.19\textwidth}
    \includegraphics[width=\linewidth]{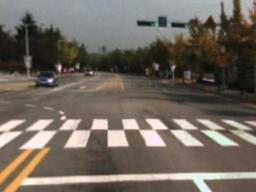}
    \caption{VI Image}
    \label{fig:4b}
  \end{subfigure}
  \hfill
  \begin{subfigure}{0.19\textwidth}
    \includegraphics[width=\linewidth]{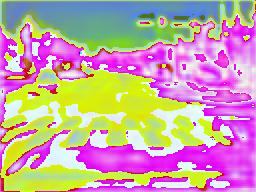}
    \caption{Epoch 10}
    \label{fig:4c}
  \end{subfigure}
  \hfill
  \begin{subfigure}{0.19\textwidth}
    \includegraphics[width=\linewidth]{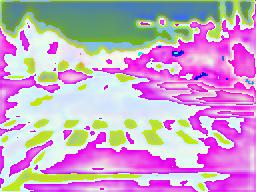}
    \caption{Epoch 20}
    \label{fig:4d}
  \end{subfigure}
  \hfill
  \begin{subfigure}{0.19\textwidth}
    \includegraphics[width=\linewidth]{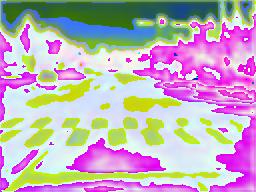}
    \caption{Epoch 30}
    \label{fig:4e}
  \end{subfigure}

  \vspace{10pt}

  \begin{subfigure}{0.19\textwidth}
    \includegraphics[width=\linewidth]{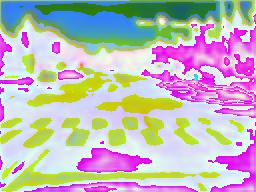}
    \caption{Epoch 40}
    \label{fig:4f}
  \end{subfigure}
  \hfill
  \begin{subfigure}{0.19\textwidth}
    \includegraphics[width=\linewidth]{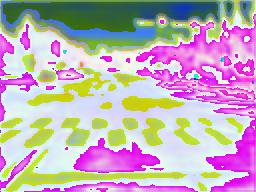}
    \caption{Epoch 50}
    \label{fig:4g}
  \end{subfigure}
  \hfill
  \begin{subfigure}{0.19\textwidth}
    \includegraphics[width=\linewidth]{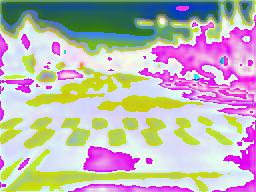}
    \caption{Epoch 60}
    \label{fig:4h}
  \end{subfigure}
  \hfill
  \begin{subfigure}{0.19\textwidth}
    \includegraphics[width=\linewidth]{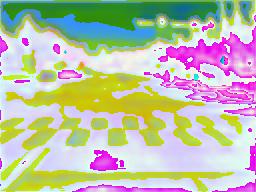}
    \caption{Epoch 70}
    \label{fig:4i}
  \end{subfigure}
  \hfill
  \begin{subfigure}{0.19\textwidth}
    \includegraphics[width=\linewidth]{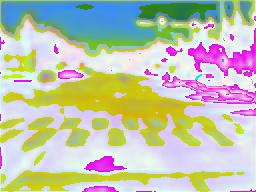}
    \caption{Epoch 80}
    \label{fig:4j}
  \end{subfigure}

  \caption{Handle module output on the training set.}
  \label{fig:4}
\end{figure}

\begin{figure}[h]
  \centering

  \begin{subfigure}{0.19\textwidth}
    \includegraphics[width=\linewidth]{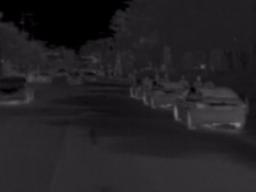}
    \caption{TIR Image}
    \label{fig:5a}
  \end{subfigure}
  \hfill
  \begin{subfigure}{0.19\textwidth}
    \includegraphics[width=\linewidth]{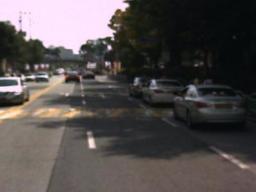}
    \caption{VI Image}
    \label{fig:5b}
  \end{subfigure}
  \hfill
  \begin{subfigure}{0.19\textwidth}
    \includegraphics[width=\linewidth]{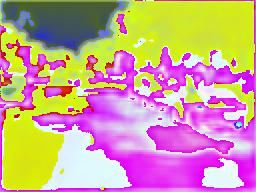}
    \caption{Epoch 10}
    \label{fig:5c}
  \end{subfigure}
  \hfill
  \begin{subfigure}{0.19\textwidth}
    \includegraphics[width=\linewidth]{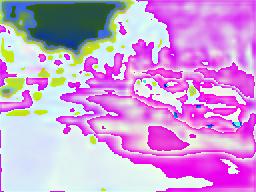}
    \caption{Epoch 20}
    \label{fig:5d}
  \end{subfigure}
  \hfill
  \begin{subfigure}{0.19\textwidth}
    \includegraphics[width=\linewidth]{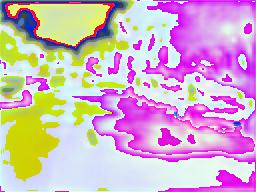}
    \caption{Epoch 30}
    \label{fig:5e}
  \end{subfigure}

  \vspace{10pt}

  \begin{subfigure}{0.19\textwidth}
    \includegraphics[width=\linewidth]{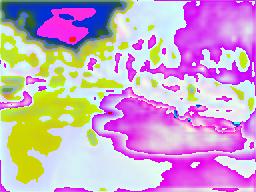}
    \caption{Epoch 40}
    \label{fig:5f}
  \end{subfigure}
  \hfill
  \begin{subfigure}{0.19\textwidth}
    \includegraphics[width=\linewidth]{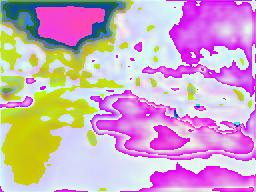}
    \caption{Epoch 50}
    \label{fig:5g}
  \end{subfigure}
  \hfill
  \begin{subfigure}{0.19\textwidth}
    \includegraphics[width=\linewidth]{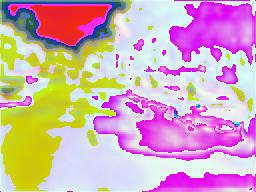}
    \caption{Epoch 60}
    \label{fig:5h}
  \end{subfigure}
  \hfill
  \begin{subfigure}{0.19\textwidth}
    \includegraphics[width=\linewidth]{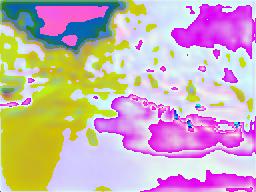}
    \caption{Epoch 70}
    \label{fig:5i}
  \end{subfigure}
  \hfill
  \begin{subfigure}{0.19\textwidth}
    \includegraphics[width=\linewidth]{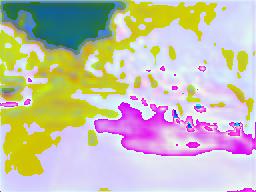}
    \caption{Epoch 80}
    \label{fig:5j}
  \end{subfigure}

  \caption{Handle module output on the testing set.}
  \label{fig:5}
\end{figure}

Fundamentally, the Handle module serves as an abstract semantic construction network, utilizing a more intricate structure to build a more abstract semantic space, consequently enhancing the Bowl module's understanding of mapping relationships. Traditional U-net structures primarily construct semantic spaces through a single encoder. Even with ongoing enhancements to the convolutional block's feature extraction capabilities, it remains challenging to match the capabilities of a complete U-net. Thus, we consider the role of the Handle module within LadleNet as the construction of an abstract semantic space. To validate this conclusion, we provide output from LadleNet, showcasing features from the Handle module during various epochs of training, as depicted in Figures \ref{fig:4} and \ref{fig:5}.

\begin{figure}[h]
    \centering
    \includegraphics[width=\textwidth]{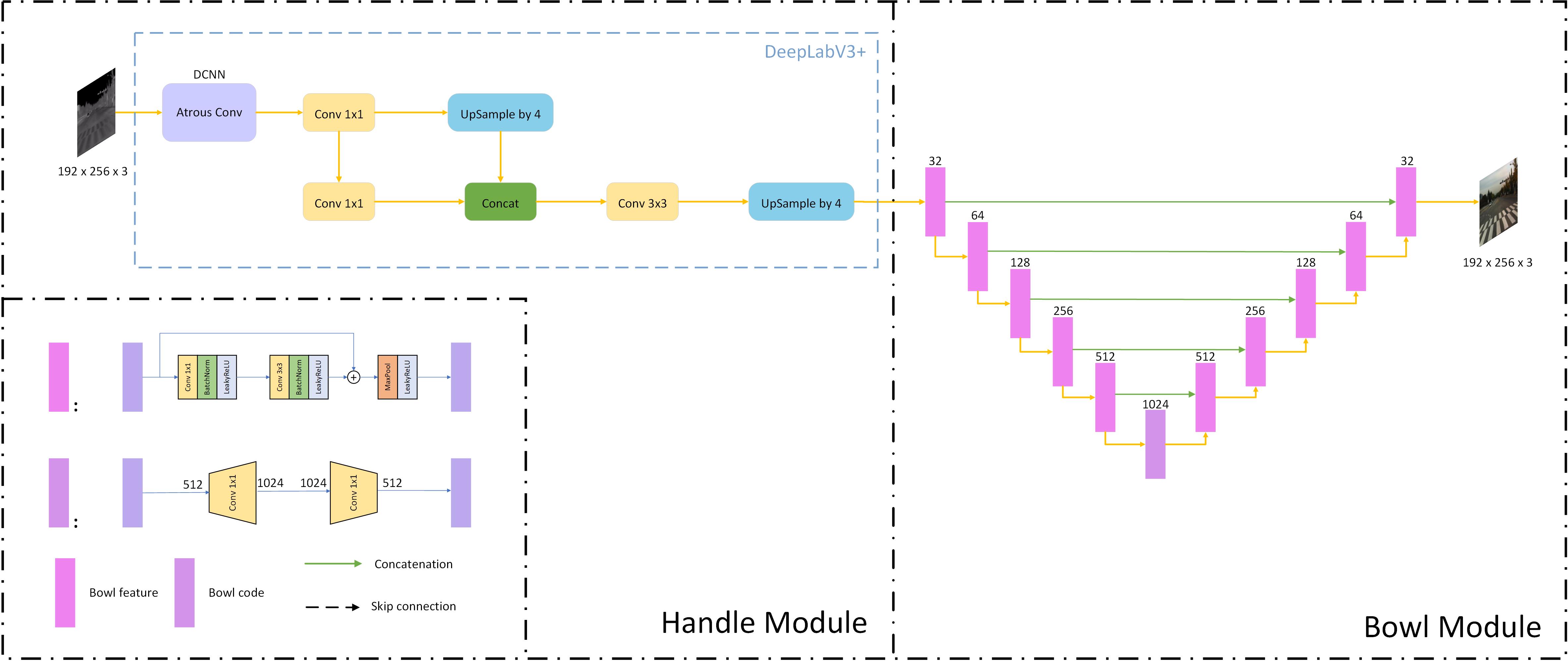}
    \caption{The architecture of the LadleNet+ model, where the Handle module uses the pre-trained DeepLabV3+ model.}
    \label{fig:6}
\end{figure}

Figures \ref{fig:4} and \ref{fig:5} illustrate that the Handle module fundamentally engages in the construction of an abstract semantic space. The transformation process exhibited by the Handle module's output in these figures resembles the scene segmentation process observed in semantic segmentation applications. This resemblance led us to the notion that employing a pre-trained semantic segmentation model as the backbone network for the Handle module could substantially enhance model performance. To address this, we introduce the LadleNet+ model, which replaces the original U-net within LadleNet with the DeepLabV3+ model trained on the street scene dataset Cityscapes\cite{net_3}. The architectural depiction of the LadleNet+ model can be seen in Figure \ref{fig:6}.

Incorporating the pre-trained DeepLabV3+ model as the backbone network for the Handle module further facilitates the construction of a more abstract semantic space, thereby enhancing the accuracy and authenticity of the resulting mapping relationships in the translated VI images. However, due to the disparity in feature map sizes between DeepLabV3+ layers and the Bowl module, the feature aggregation and skip-connection structures between the Handle module and the Bowl module were discarded when implementing LadleNet+. Moreover, given that existing publicly available pre-trained DeepLabV3+ models are based on VI images, the semantic space construction capacity for TIR images is somewhat limited. Nonetheless, despite these minor limitations, utilizing the pre-trained DeepLabV3+ as the backbone network for the Handle module significantly enhances the abstraction level of the semantic space.

The scalability of LadleNet primarily manifests in the choice of the backbone network for the Handle module. The approach presented in this paper involves replacing the Handle module's backbone network in the basic LadleNet with the pre-trained DeepLabV3+ model, yielding the enhanced performance of LadleNet+. In practice, the backbone network for the Handle module (HandleNet) can also be substituted with other networks possessing robust semantic modeling capabilities based on task requirements, including networks like Mask R-CNN\cite{net_5}, PSPNet\cite{net_6}, DANet\cite{net_7}, among others. We believe that the true strength of LadleNet lies in this versatility, whereby the problem of enhancing image translation performance is transformed into selecting a suitable semantic segmentation network, thus avoiding ineffective and directionless improvement endeavors.

\subsubsection*{Bowl Module}
The Bowl module consists of a 5-layer U-net, where each convolutional block comprises two convolutional layers with a kernel size of 3x3. Between every two consecutive convolutional layers within a block, a skip connection is introduced, followed by a 1x1 convolutional layer to reduce the feature map size by half. The primary function of the Bowl module within the U-net is to decode the high-level semantic information from the Handle module, thus generating the final mapped VI image. The unique characteristic of the U-net in the Bowl module lies in its ability to aggregate features efficiently through concatenation, which is vital for semantic space decoding. Additionally, this U-net maintains skip connections with high-level features from the Handle module to prevent any degradation resulting from feature aggregation.

Regarding the network within the Bowl module, we believe that employing the aforementioned U-net with feature aggregation and skip connection mechanisms is sufficient. On one hand, the crux of LadleNet hinges on the construction of the semantic space in the Handle module, as only a sufficiently abstract semantic space can better capture the mapping from TIR images to VI images. On the other hand, the feature aggregation and skip connection mechanisms we introduced to the U-net significantly enhance the decoding capability of the Bowl module. Introducing excessive complexity to the network in the Bowl module might lead to potential model degradation. However, the effectiveness of extending the Bowl module warrants further in-depth investigation through specific experiments. Nonetheless, we assert that, at this stage, the expansion of the Handle module remains of greater importance.

\subsection{Loss Function}
LadleNet aims to achieve the highest quality of image translation results, and the generated images should appear as realistic as possible. To achieve this goal, we have chosen a weighted combination of L1 loss and MS-SSIM loss as the loss function for LadleNet. The L1 loss measures pixel-level image dissimilarity, enabling the generated images to capture more fine details. On the other hand, the MS-SSIM loss assesses the SSIM loss at different scales, thereby providing better assessment of both local and global visual aspects. This contributes to a more realistic overall appearance of the generated images. By assigning different weights to the L1 loss and the MS-SSIM loss, we achieve more efficient loss evaluation.

\subsubsection*{L1 Loss}
The L1 loss is one of the most commonly used loss functions in image generation and image translation tasks. Compared to the L2 loss, the L1 loss tends to yield clearer generated results. In LadleNet, the L1 loss is employed to measure the pixel-level differences between the generated fake VI (FVI) images obtained from input TIR images and the actual VI images. The L1 loss between the generated VI image and the true VI image is defined as follows:
\begin{align}
\mathscr{L}_{l1} &= \frac{1}{H \times W \times 3} \quad \sum^H_{h=1}  \sum^W_{w=1} \sum^3_{c=1} \quad \lvert I_{FVI}^{(h,w,c)} - I_{VI}^{(h,w,c)} \rvert \tag{1}
\end{align}

Where, $I_{TIR} \in \mathscr{R}^{H \times W \times 3}$ represents the input TIR image, $I_{FVI} \in \mathscr{R}^{H \times W \times 3}$ represents the generated fake VI image obtained from LadleNet, and $I_{VI} \in \mathscr{R}^{H \times W \times 3}$ represents the actual VI image.

\subsubsection*{MS-SSIM Loss}
In the field of image generation and translation, the Structural Similarity Index (SSIM) \cite{net_8} is also a commonly used metric for assessing image quality. SSIM evaluates image quality from the perspective of human visual perception by measuring structural information differences between images. The expression of this index is as follows:
\begin{align}
SSIM_s(I_{\text{FVI}}, I_{\text{VI}}) = \frac{{(2\mu_{I_{\text{FVI}}} \mu_{I_{\text{VI}}} + C_1) \cdot (2\sigma_{I_{\text{FVI}},I_{\text{VI}}} + C_2)}}{{(\mu_{I_{\text{FVI}}}^2 + \mu_{I_{\text{VI}}}^2 + C_1) \cdot (\sigma_{I_{\text{FVI}}}^2 + \sigma_{I_{\text{VI}}}^2 + C_2)}} \tag{2}
\end{align}

Where, $s$ represents the scale, indicating the image scaling factor. $\mu_{I_{\text{FVI}}}$ and $\mu_{I_{\text{VI}}}$ are the average values of $I_{FVI}$ and $I_{VI}$ at scale $s$, respectively. $\sigma_{I_{\text{FVI}}}$ and $\sigma_{I_{\text{VI}}}$ are the standard deviations of $I_{FVI}$ and $I_{VI}$, respectively. $\sigma_{I_{\text{FVI}},I_{\text{VI}}}$ represents the covariance of $I_{FVI}$ and $I_{VI}$ at scale $s$. $C_1$ and $C_2$ are constants introduced to stabilize the calculation.

While SSIM is effective in supervising the quality of generated images, it is limited to supervising images at a single scale, leading to potential loss of details in the generated images. The Multiscale Structural Similarity (MS-SSIM) Index proposed by Wang et al. \cite{net_9} measures SSIM values across multiple scales, allowing for better supervision of the model's generation performance on image details. Its mathematical expression is as follows:
\begin{align}
\text{MS-SSIM}(I_{\text{FVI}}, I_{\text{VI}}) = \sum_{s=1}^{S} 	\omega_s  \cdot SSIM_s(I_{\text{FVI}}, I_{\text{VI}}) \tag{3}
\end{align}

Where, $S$ represents the total number of scales, and $\omega_s$ represents the weight associated with each scale. When evaluating the MS-SSIM index between images, the goal is to maximize it; however, during model training, the loss function needs to be minimized. Therefore, to convert the MS-SSIM index into a loss function, the MS-SSIM loss function is constructed as follows:
\begin{align}
\mathscr{L}_{MS-SSIM}=1-\text{MS-SSIM}(I_{FVI},I_{VI}) \tag{4}
\end{align}

By converting the $MS-SSIM$ index into the loss function $\mathscr{L}_{MS-SSIM}$, the final generation results can be supervised by minimizing this loss function, thus enhancing the quality of the generated images by the model.

\subsubsection*{Total Loss}
In order to maximize the supervisory effects of $\mathscr{L}{l1}$ and $\mathscr{L}{MS-SSIM}$, we attempt to combine these two loss functions with appropriate weights to create a more powerful composite loss function. The total loss is defined as follows:
\begin{align}
\mathscr{L}_{Total}=\alpha \cdot \mathscr{L}_{MS-SSIM}  + (1-\alpha) \cdot \mathscr{L}_{l1} \tag{5}
\end{align}

Where, $\alpha$ represents the weight controlling the proportion of different losses in the total loss. To obtain a more strongly supervised and efficient loss function, we refer to the work by Zhao et al. \cite{net_10}. In their paper, exploration of the supervisory effects of different image quality metrics is conducted. We adopt the optimal parameter settings provided in the paper, specifically setting the weights in $\mathscr{L}_{MS-SSIM}$ as $\omega = {0.5, 1, 2, 4, 8}$ and setting $\alpha=0.84$.

\section{Experiments}
In this section, we conduct experimental validations of the proposed model architecture and compare it with other models in the field. We evaluate the performance of our proposed model through both qualitative and quantitative analyses. Furthermore, comprehensive ablation experiments on the model structure are performed to ascertain its effectiveness.

\subsection{Dataset}
The KAIST dataset\cite{ex_1} is a multispectral road dataset containing 95,328 color-thermal pairs. It covers road scenes in campus, street, and rural environments, and provides coarse time periods (daytime and nighttime) as well as fine time periods (sunrise, morning, afternoon, sunset, night, and dawn). For training and validation of our model, we select Set 01 from the training set, comprising 8,035 pairs of images, and Set 07 from the test set, comprising 8,141 pairs of images, forming our experimental dataset with a total of 16,176 color-thermal pairs. Among these, 80\% are used for training and 20\% for testing, exclusively focusing on daytime scenes. To ensure fair performance comparison among different models, we train existing TIR-to-VI image translation models on the training set and evaluate their performance on the test set.

\subsection{Implementation details}
Images from the dataset are transformed into tensors and resized to 300x400 dimensions, followed by central cropping to 192x256 dimensions. Both TIR and VI images are treated as 3-channel RGB images. Both LadleNet and LadleNet+ are trained for 120 epochs with a batch size of 40 samples. The initial learning rate is set to 0.01. Whenever the loss value does not decrease for 2 consecutive epochs, the learning rate is reduced by a factor of 0.1, with no more learning rate changes occurring for the subsequent 5 epochs. Both models use the Adam optimizer, with all parameters set to default values, and $amsgrad$ is set to True. All training and testing procedures are performed on an NVIDIA A40 GPU and an AMD EPYC 7543 CPU. The average duration to train one epoch for LadleNet is 8 minutes, while for LadleNet+ it is 10 minutes. The DeepLabV3+ model used in LadleNet+ employs ResNet101\cite{net_10} as its backbone network and is pretrained on the Cityscapes dataset\cite{net_3}, a street scene dataset.

\subsection{Comparative experiment}
We conducted comparisons between our proposed LadleNet and LadleNet+ models and existing methods for TIR-to-VI image translation, as well as some foundational image generation benchmark models. The compared methods include TIR2Lab\cite{ex_2}, U-net, U-net\_IR2VI\cite{ex_3}, Pix2Pix, and Pix2Pix\_IR2VI\cite{ex_4}. While code for some of these methods might not be publicly available, we followed the descriptions in the respective papers to replicate the models\footnote{The replication code for U-net\_IR2VI and Pix2Pix\_IR2VI will be made available at \url{https://github.com/Ach-1914/LadleNet/tree/main/Model/}}. To ensure fair quantitative comparisons among different models, we trained all models on the same training set and evaluated their performance on the same test set. We adopted four metrics to measure the image discrepancies, including Structural Similarity Index (SSIM), Multiscale Structural Similarity Index (MS-SSIM), L1 metric, and Peak Signal-to-Noise Ratio (PSNR). These metrics effectively quantify the disparities between the generated VI images and the ground truth VI images, and they are commonly used metrics in the field of image translation. The comparative results of various models on the test set are presented in Table \ref{tab:1}.

\begin{figure}[h]
    \centering
    \includegraphics[width=\textwidth]{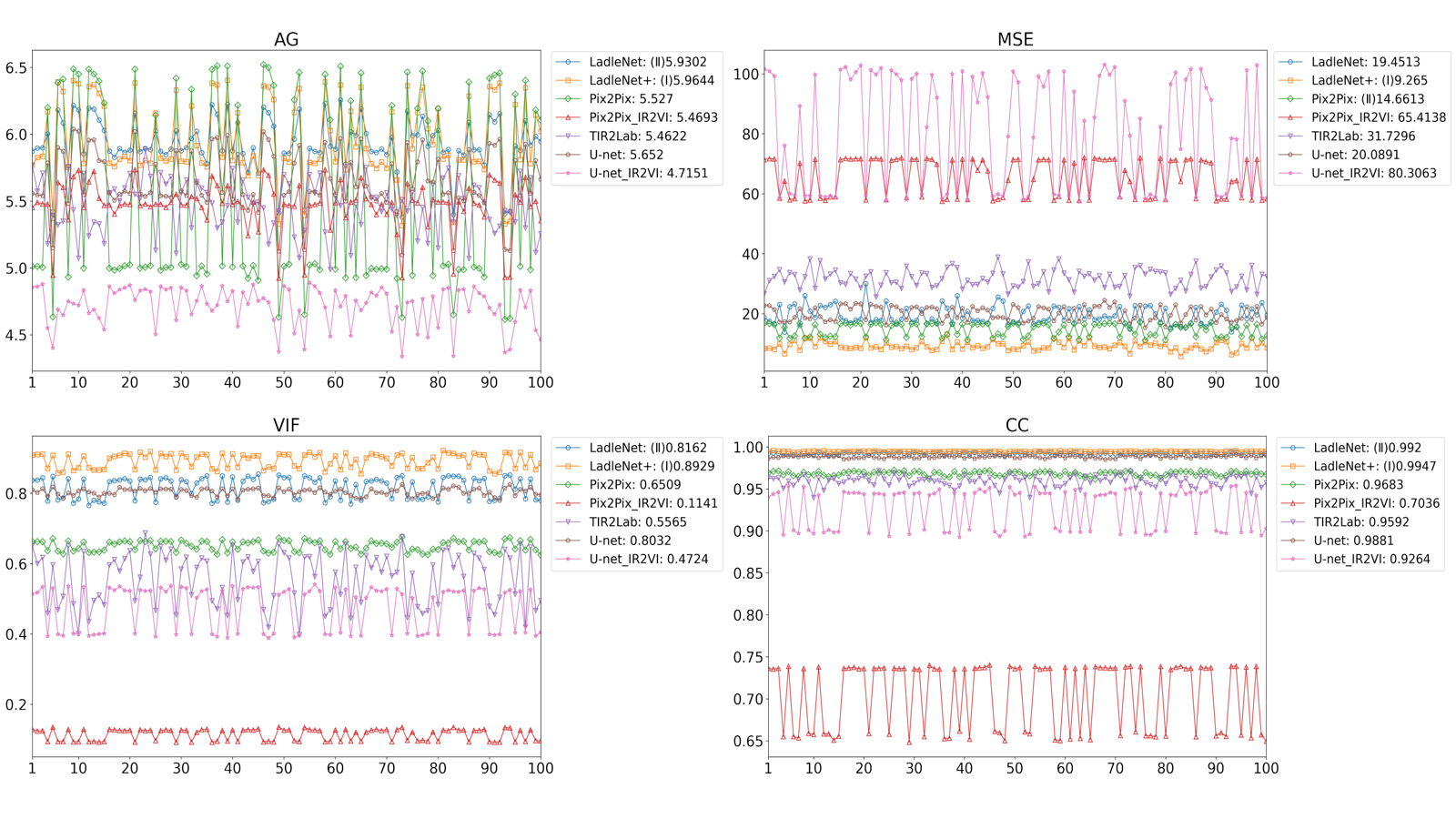}
    \caption{Comparison of the four metrics, namely AG, MSE, VIF, and CC, for each model across 100 image pairs. The numerical values in the legend represent the average metric scores for the 100 image pairs per model, with the top-performing result labeled as \uppercase\expandafter{\romannumeral1}, and the second-best performance labeled as \uppercase\expandafter{\romannumeral2}.}
    \label{fig:8}
\end{figure}

\begin{table}[htbp]
    \centering
    \caption{Average metrics results for each model on the test set (red indicates the best results, blue indicates the second best results).}
    \begin{tabular}{@{}lllll@{}}
        \toprule
        \quad & SSIM & MS-SSIM & L1 & PSNR \\
        \midrule
        TIR2Lab & 0.6816 & 0.1998 & 0.0525 & 21.5512 \\
        U-net & 0.7619 & 0.4268 & 0.0515 & 23.4234 \\
        U-net\_IR2VI & 0.8060 & 0.2792 & 0.0412 & 23.3618 \\
        Pix2Pix & 0.6476 & 0.1611 & 0.0719 & 18.9515 \\
        Pix2Pix\_IR2VI & 0.7398 & 0.1802 & 0.0636 & 20.4982 \\
        LadleNet & \textcolor{blue}{0.8518} & \textcolor{blue}{0.6292}  & \textcolor{blue}{0.0297} & \textcolor{blue}{26.6475} \\
        LadleNet+ & \textcolor{red}{0.8798} & \textcolor{red}{0.7561} & \textcolor{red}{0.0223} & \textcolor{red}{27.5494} \\
        
        \bottomrule
    \end{tabular}
    \label{tab:1}
\end{table}

\begin{figure}[htbp]
\centering
\begin{minipage}[c]{0.1\textwidth}
  \vspace{\fill}
  (a)
  \label{fig:7a}
\end{minipage}%
\begin{minipage}[c]{0.17\textwidth}
  \vspace{\fill}
  \includegraphics[width=\linewidth]{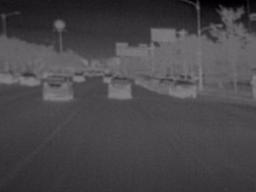}
\end{minipage}%
\begin{minipage}[c]{0.17\textwidth}
  \vspace{\fill}
  \includegraphics[width=\linewidth]{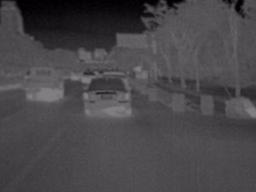}
\end{minipage}%
\begin{minipage}[c]{0.17\textwidth}
  \vspace{\fill}
  \includegraphics[width=\linewidth]{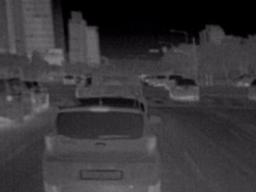}
\end{minipage}%
\begin{minipage}[c]{0.17\textwidth}
  \vspace{\fill}
  \includegraphics[width=\linewidth]{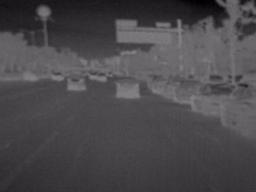}
\end{minipage}%
\begin{minipage}[c]{0.17\textwidth}
  \vspace{\fill}
  \includegraphics[width=\linewidth]{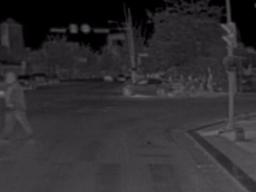}
\end{minipage}\par

\begin{minipage}[c]{0.1\textwidth}
  \vspace{\fill}
  (b)
  \label{fig:7b}
\end{minipage}%
\begin{minipage}[c]{0.17\textwidth}
  \vspace{\fill}
  \includegraphics[width=\linewidth]{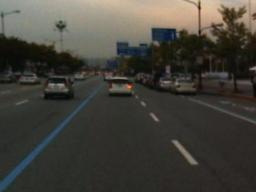}
\end{minipage}%
\begin{minipage}[c]{0.17\textwidth}
  \vspace{\fill}
  \includegraphics[width=\linewidth]{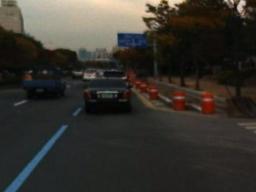}
\end{minipage}%
\begin{minipage}[c]{0.17\textwidth}
  \vspace{\fill}
  \includegraphics[width=\linewidth]{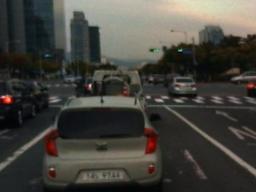}
\end{minipage}%
\begin{minipage}[c]{0.17\textwidth}
  \vspace{\fill}
  \includegraphics[width=\linewidth]{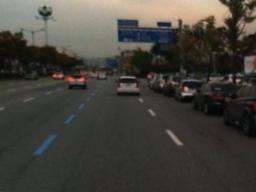}
\end{minipage}%
\begin{minipage}[c]{0.17\textwidth}
  \vspace{\fill}
  \includegraphics[width=\linewidth]{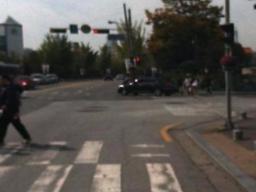}
\end{minipage}\par

\begin{minipage}[c]{0.1\textwidth}
  \vspace{\fill}
  (c)
  \label{fig:7c}
\end{minipage}%
\begin{minipage}[c]{0.17\textwidth}
  \vspace{\fill}
  \includegraphics[width=\linewidth]{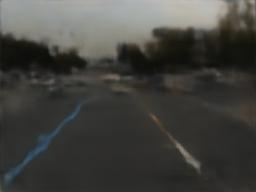}
\end{minipage}%
\begin{minipage}[c]{0.17\textwidth}
  \vspace{\fill}
  \includegraphics[width=\linewidth]{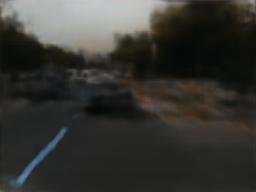}
\end{minipage}%
\begin{minipage}[c]{0.17\textwidth}
  \vspace{\fill}
  \includegraphics[width=\linewidth]{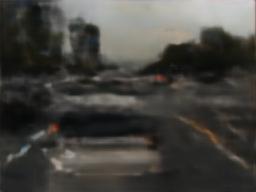}
\end{minipage}%
\begin{minipage}[c]{0.17\textwidth}
  \vspace{\fill}
  \includegraphics[width=\linewidth]{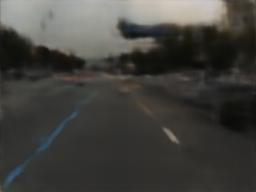}
\end{minipage}%
\begin{minipage}[c]{0.17\textwidth}
  \vspace{\fill}
  \includegraphics[width=\linewidth]{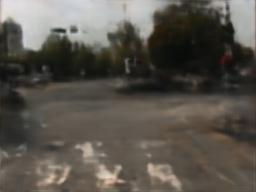}
\end{minipage}\par

\begin{minipage}[c]{0.1\textwidth}
  \vspace{\fill}
  (d)
  \label{fig:7d}
\end{minipage}%
\begin{minipage}[c]{0.17\textwidth}
  \vspace{\fill}
  \includegraphics[width=\linewidth]{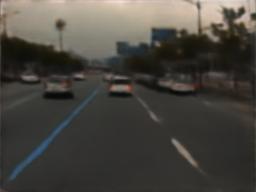}
\end{minipage}%
\begin{minipage}[c]{0.17\textwidth}
  \vspace{\fill}
  \includegraphics[width=\linewidth]{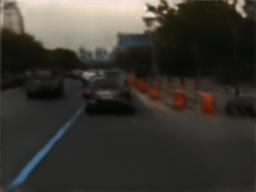}
\end{minipage}%
\begin{minipage}[c]{0.17\textwidth}
  \vspace{\fill}
  \includegraphics[width=\linewidth]{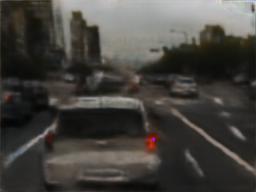}
\end{minipage}%
\begin{minipage}[c]{0.17\textwidth}
  \vspace{\fill}
  \includegraphics[width=\linewidth]{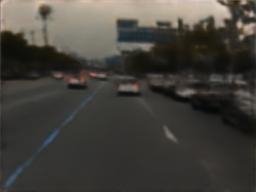}
\end{minipage}%
\begin{minipage}[c]{0.17\textwidth}
  \vspace{\fill}
  \includegraphics[width=\linewidth]{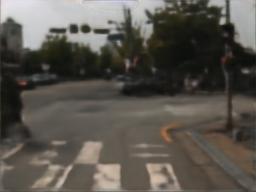}
\end{minipage}\par

\begin{minipage}[c]{0.1\textwidth}
  \vspace{\fill}
  (e)
  \label{fig:7e}
\end{minipage}%
\begin{minipage}[c]{0.17\textwidth}
  \vspace{\fill}
  \includegraphics[width=\linewidth]{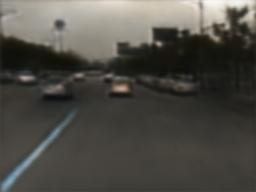}
\end{minipage}%
\begin{minipage}[c]{0.17\textwidth}
  \vspace{\fill}
  \includegraphics[width=\linewidth]{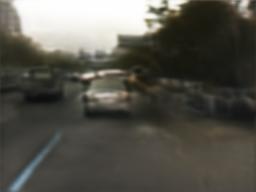}
\end{minipage}%
\begin{minipage}[c]{0.17\textwidth}
  \vspace{\fill}
  \includegraphics[width=\linewidth]{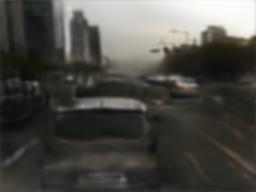}
\end{minipage}%
\begin{minipage}[c]{0.17\textwidth}
  \vspace{\fill}
  \includegraphics[width=\linewidth]{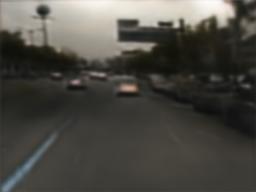}
\end{minipage}%
\begin{minipage}[c]{0.17\textwidth}
  \vspace{\fill}
  \includegraphics[width=\linewidth]{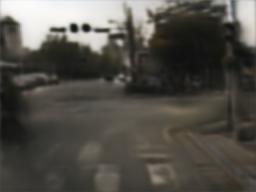}
\end{minipage}\par

\begin{minipage}[c]{0.1\textwidth}
  \vspace{\fill}
  (f)
  \label{fig:7f}
\end{minipage}%
\begin{minipage}[c]{0.17\textwidth}
  \vspace{\fill}
  \includegraphics[width=\linewidth]{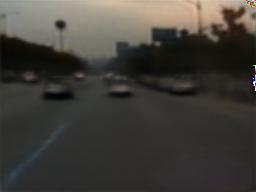}
\end{minipage}%
\begin{minipage}[c]{0.17\textwidth}
  \vspace{\fill}
  \includegraphics[width=\linewidth]{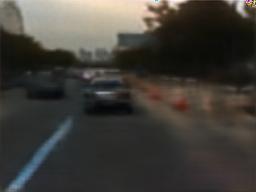}
\end{minipage}%
\begin{minipage}[c]{0.17\textwidth}
  \vspace{\fill}
  \includegraphics[width=\linewidth]{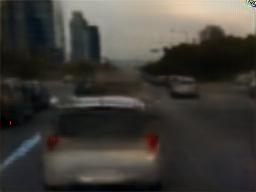}
\end{minipage}%
\begin{minipage}[c]{0.17\textwidth}
  \vspace{\fill}
  \includegraphics[width=\linewidth]{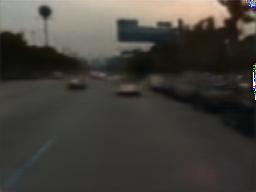}
\end{minipage}%
\begin{minipage}[c]{0.17\textwidth}
  \vspace{\fill}
  \includegraphics[width=\linewidth]{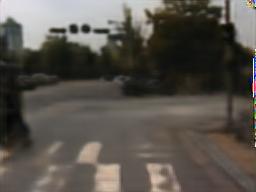}
\end{minipage}\par

\begin{minipage}[c]{0.1\textwidth}
  \vspace{\fill}
  (g)
  \label{fig:7g}
\end{minipage}%
\begin{minipage}[c]{0.17\textwidth}
  \vspace{\fill}
  \includegraphics[width=\linewidth]{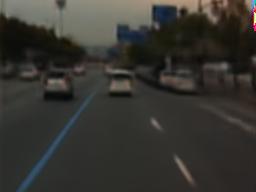}
\end{minipage}%
\begin{minipage}[c]{0.17\textwidth}
  \vspace{\fill}
  \includegraphics[width=\linewidth]{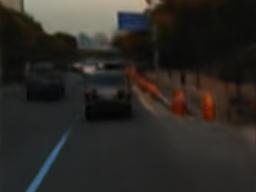}
\end{minipage}%
\begin{minipage}[c]{0.17\textwidth}
  \vspace{\fill}
  \includegraphics[width=\linewidth]{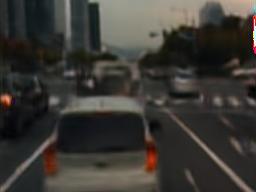}
\end{minipage}%
\begin{minipage}[c]{0.17\textwidth}
  \vspace{\fill}
  \includegraphics[width=\linewidth]{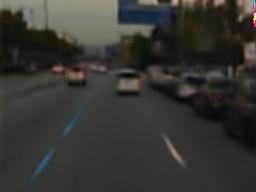}
\end{minipage}%
\begin{minipage}[c]{0.17\textwidth}
  \vspace{\fill}
  \includegraphics[width=\linewidth]{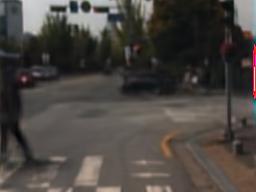}
\end{minipage}\par

\begin{minipage}[c]{0.1\textwidth}
  \vspace{\fill}
  (h)
  \label{fig:7h}
\end{minipage}%
\begin{minipage}[c]{0.17\textwidth}
  \vspace{\fill}
  \includegraphics[width=\linewidth]{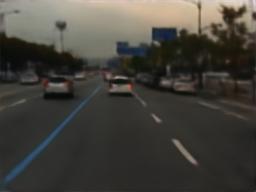}
\end{minipage}%
\begin{minipage}[c]{0.17\textwidth}
  \vspace{\fill}
  \includegraphics[width=\linewidth]{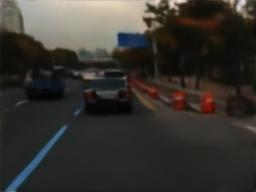}
\end{minipage}%
\begin{minipage}[c]{0.17\textwidth}
  \vspace{\fill}
  \includegraphics[width=\linewidth]{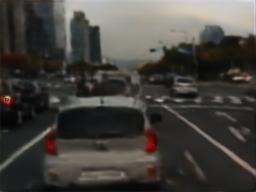}
\end{minipage}%
\begin{minipage}[c]{0.17\textwidth}
  \vspace{\fill}
  \includegraphics[width=\linewidth]{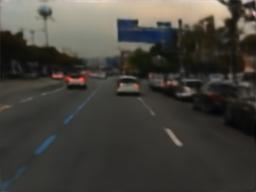}
\end{minipage}%
\begin{minipage}[c]{0.17\textwidth}
  \vspace{\fill}
  \includegraphics[width=\linewidth]{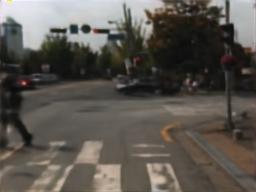}
\end{minipage}\par

\begin{minipage}[c]{0.1\textwidth}
  \vspace{\fill}
  (i)
  \label{fig:7i}
\end{minipage}%
\begin{minipage}[c]{0.17\textwidth}
  \vspace{\fill}
  \includegraphics[width=\linewidth]{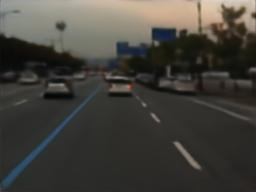}
\end{minipage}%
\begin{minipage}[c]{0.17\textwidth}
  \vspace{\fill}
  \includegraphics[width=\linewidth]{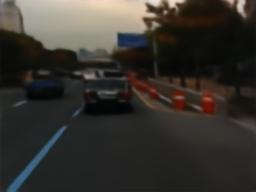}
\end{minipage}%
\begin{minipage}[c]{0.17\textwidth}
  \vspace{\fill}
  \includegraphics[width=\linewidth]{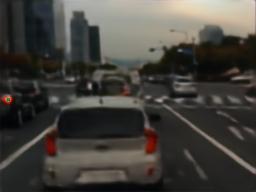}
\end{minipage}%
\begin{minipage}[c]{0.17\textwidth}
  \vspace{\fill}
  \includegraphics[width=\linewidth]{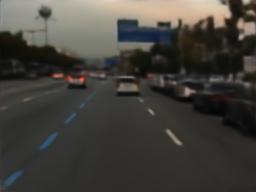}
\end{minipage}%
\begin{minipage}[c]{0.17\textwidth}
  \vspace{\fill}
  \includegraphics[width=\linewidth]{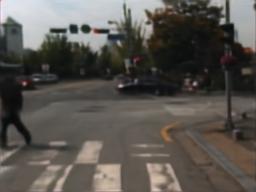}
\end{minipage}\par

\caption{Comparison of translation results for 5 typical TIR images on the test set. (a) TIR images. (b) VI images, i.e., ground truth. (c) TIR2Lab. (d) U-net. (e) U-net\_IR2VI. (f) Pix2Pix. (g) Pix2Pix\_IR2VI. (h) LadleNet. (i) LadleNet+.}
\label{fig:7}
\end{figure}

As shown in Table \ref{tab:1}, both LadleNet and LadleNet+ achieve state-of-the-art performance across the four metrics, particularly excelling in the MS-SSIM metric. On one hand, the incorporation of MS-SSIM as part of the loss function significantly enhances the model's performance in this particular metric. On the other hand, the improvement in the MS-SSIM metric also has a positive impact on the model's performance in terms of SSIM and PSNR metrics, resulting in state-of-the-art performance across all metrics. In order to provide a more intuitive visualization of the final translation results of the TIR images obtained by various models, we present a comparison of the output results of each model in Figure \ref{fig:7}.

In Fig \ref{fig:7}, the first two rows display the original TIR images and corresponding VI images, while the last two rows depict the output images of LadleNet and LadleNet+. From the results, it is evident that the images generated by LadleNet and LadleNet+ are the most realistic and exhibit the highest clarity. The outcomes of the other models all exhibit certain deficiencies. For instance, TIR2Lab (Figure \ref{fig:7}c), U-net\_IR2VI (Figure \ref{fig:7}e), and Pix2Pix (Figure \ref{fig:7}f) translations show distortions, failing to accurately map the TIR images. In contrast, U-net (Figure \ref{fig:7}d) and Pix2Pix\_IR2VI (Figure \ref{fig:7}g) yield output that correctly reflects the scene's structure and color information, albeit with reduced clarity.

While LadleNet and LadleNet+ exhibit favorable results in qualitative comparisons, this does not imply that the generated image quality is optimal. There is still room for improvement in terms of image clarity and edge details in the images produced by LadleNet and LadleNet+. Qualitative analysis alone struggles to objectively describe the distinctions between the outputs of different models. To conduct a fairer comparison of image quality, we randomly selected 100 color-thermal pairs from the test set for quantitative experiments. We utilized four metrics, including AG (Artifacts Grade), MSE (Mean Squared Error), VIF (Visual Information Fidelity), and CC (Correlation Coefficient), to evaluate the discrepancy in quality between the VI images obtained after the thermal-to-visible translation by various models and the ground truth VI images. The results of the measurements for the 100 VI images produced by different models using these four metrics are presented in Figure \ref{fig:8}.

In the quantitative assessment across 100 color-thermal pairs, our proposed LadleNet+ and LadleNet respectively secured the first and second ranks in the AG, VIF, and CC metrics. This outcome underscores that the images translated by LadleNet+ and LadleNet exhibit superior visual perceptual quality and higher clarity. In terms of the MSE metric, LadleNet+ took the first position, with Pix2Pix following closely. Meanwhile, LadleNet trailed Pix2Pix in this metric. This indicates that LadleNet+ produces images of overall high quality, while the specific training approach of Pix2Pix leads to better performance in the MSE metric. Additionally, this result indirectly reinforces the effectiveness of LadleNet+ based on a pre-trained semantic segmentation model.

\subsection{Ablation experiment}
To further validate the effectiveness of the proposed method, we conducted multiple ablation experiments on the KAIST dataset. These experiments encompassed evaluating the optimization effect of LadleNet, assessing the individual components of LadleNet, and evaluating the effectiveness of pre-trained semantic segmentation in LadleNet+. In these experiments, each model was trained on the test set for 50 epochs under identical conditions, and the model's performance and convergence speed of loss values were compared.

\begin{figure}[h]
    \centering
    \includegraphics[width=0.8\textwidth]{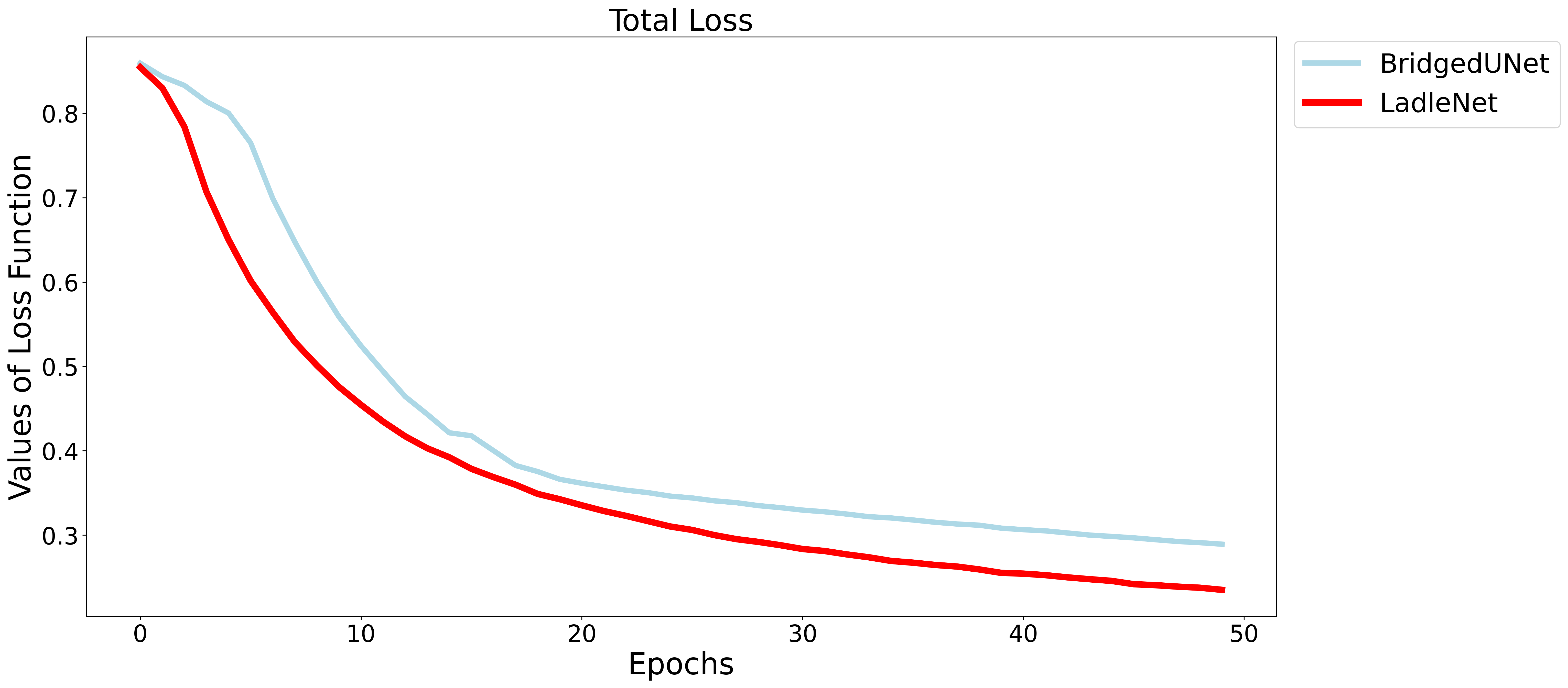}
    \caption{Comparison of LadleNet and Bridged U-net loss value convergence speeds.}
    \label{fig:9}
\end{figure}

LadleNet is an improvement over the Bridged U-net, which is not openly available to our knowledge. Therefore, we reproduced the Bridged U-net model following the methods outlined in the article\footnote{The model reproduction code for Bridged U-net will be made available at \url{https://github.com/Ach-1914/LadleNet/tree/main/Model/Bridged_U-net/}}. To ascertain the efficacy of the proposed improvements, we conducted ablation experiments on these two models. Table \ref{tab:2} presents the results of these ablation experiments, where LadleNet outperforms Bridged U-net across various metrics. Figure \ref{fig:9} illustrates the convergence speed of the loss functions for both models, with LadleNet demonstrating faster convergence and smoother overall behavior. This ablation experiment provides preliminary validation that the optimized LadleNet exhibits improved performance.

\begin{table}[htbp]
    \centering
    \caption{Comparison of model performance between LadleNet and Bridged U-net.}
    \begin{tabular}{@{}lllll@{}}
        \toprule
        \quad & SSIM & MS-SSIM & L1 & PSNR \\
        \midrule
        Bridged U-net & 0.8164 & 0.5800 & 0.0389 & 25.6045 \\
        LadleNet & 0.8493 & 0.6456  & 0.0296 & 26.8124 \\
        \bottomrule
    \end{tabular}
    \label{tab:2}
\end{table}

\begin{figure}[h]
    \centering
    \includegraphics[width=0.8\textwidth]{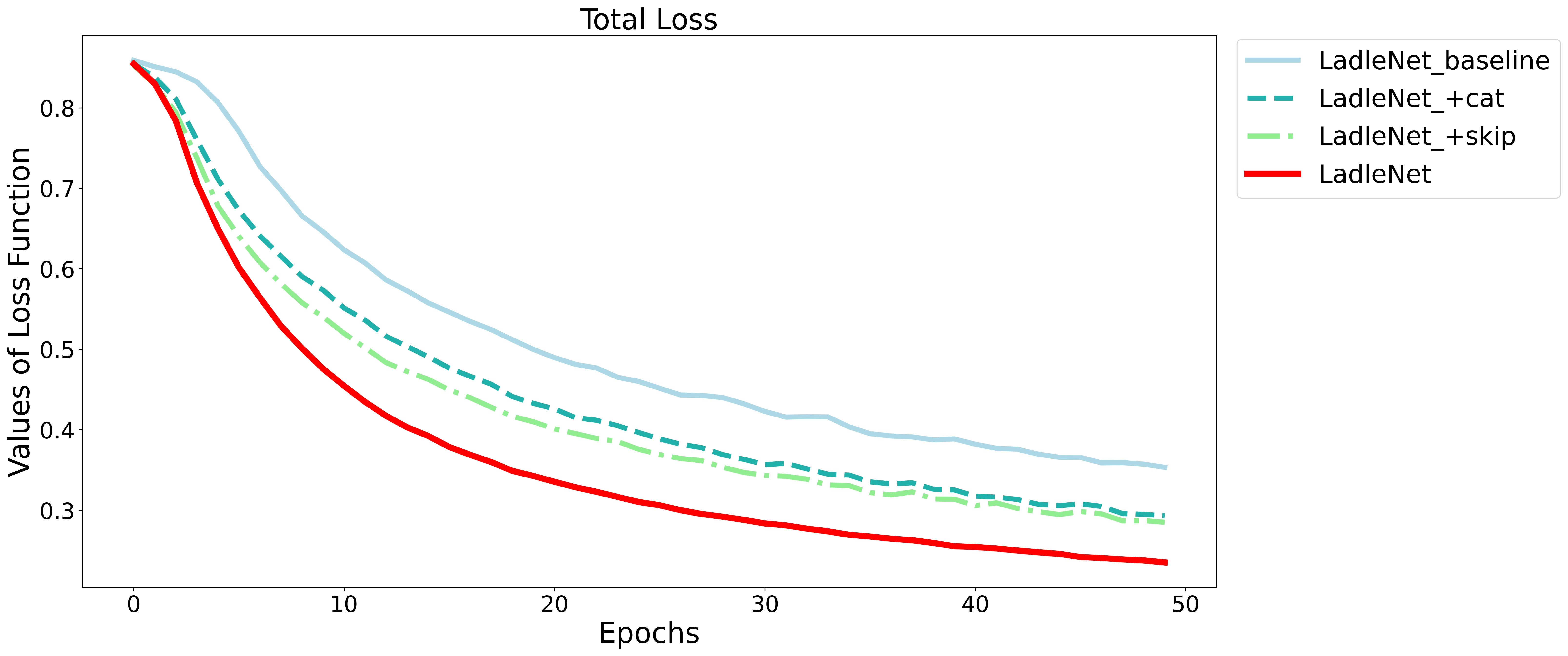}
    \caption{Comparison results of different optimization methods for the speed of convergence of model loss values.}
    \label{fig:10}
\end{figure}

To gain deeper insights into the effects of various optimization methods on LadleNet's performance, we conducted multiple sets of experiments for comparison. These experiments included a baseline model with no additional structures, LadleNet\_+skip model with only skip connections, LadleNet\_+concat model with only adjusted aggregation, and the original LadleNet model. Table \ref{tab:3} summarizes the performance of the models with different optimization methods compared to the unoptimized baseline model. Notably, the optimization method of adding only skip connections outperforms the method of adding feature aggregation alone compared to the baseline model. We believe this might be related to the structural characteristics of LadleNet. The optimization method of feature aggregation involves aggregating high-level features from the Handle module with low-level features in the Bowl module. However, these high-level features themselves are derived from aggregated low-level features. As a result, continuous feature aggregation occurs without sufficient feature extraction processes, which could explain the relatively less effective performance of this optimization method. The method of adding skip connections transfers the aggregated features from the Handle module to the feature aggregation part of the Bowl module through skip connections. This ensures the effectiveness of each feature aggregation step through the skip connections. Figure \ref{fig:10} illustrates the convergence speed variations of the loss functions for different optimization methods.

\begin{table}[htbp]
    \centering
    \caption{Comparative results of different optimization methods for model performance.}
    \begin{tabular}{@{}lllll@{}}
        \toprule
        \quad & SSIM & MS-SSIM & L1 & PSNR \\
        \midrule
        LadleNet\_baseline & 0.7878 & 0.4336 & 0.0527 & 23.3298 \\
        LadleNet\_+cat & 0.8090 & 0.4765 & 0.0417 & 24.5543 \\
        LadleNet\_+skip & 0.8265 & 0.5640 & 0.0337 & 25.9769 \\
        LadleNet & 0.8493 & 0.6456 & 0.0296 & 26.8124 \\
        \bottomrule
    \end{tabular}
    \label{tab:3}
\end{table}

\begin{table}[htbp]
    \centering
    \caption{Comparative results of pre-training methods for model performance.}
    \begin{tabular}{@{}lllll@{}}
        \toprule
        \quad & SSIM & MS-SSIM & L1 & PSNR \\
        \midrule
        LadleNet+\_no\_pre & 0.8231 & 0.6950 & 0.0333 & 24.5564 \\
        LadleNet+ & 0.8596 & 0.6283 & 0.0264 & 26.9369 \\
        \bottomrule
    \end{tabular}
    \label{tab:4}
\end{table}

\begin{figure}[h]
    \centering
    \includegraphics[width=0.8\textwidth]{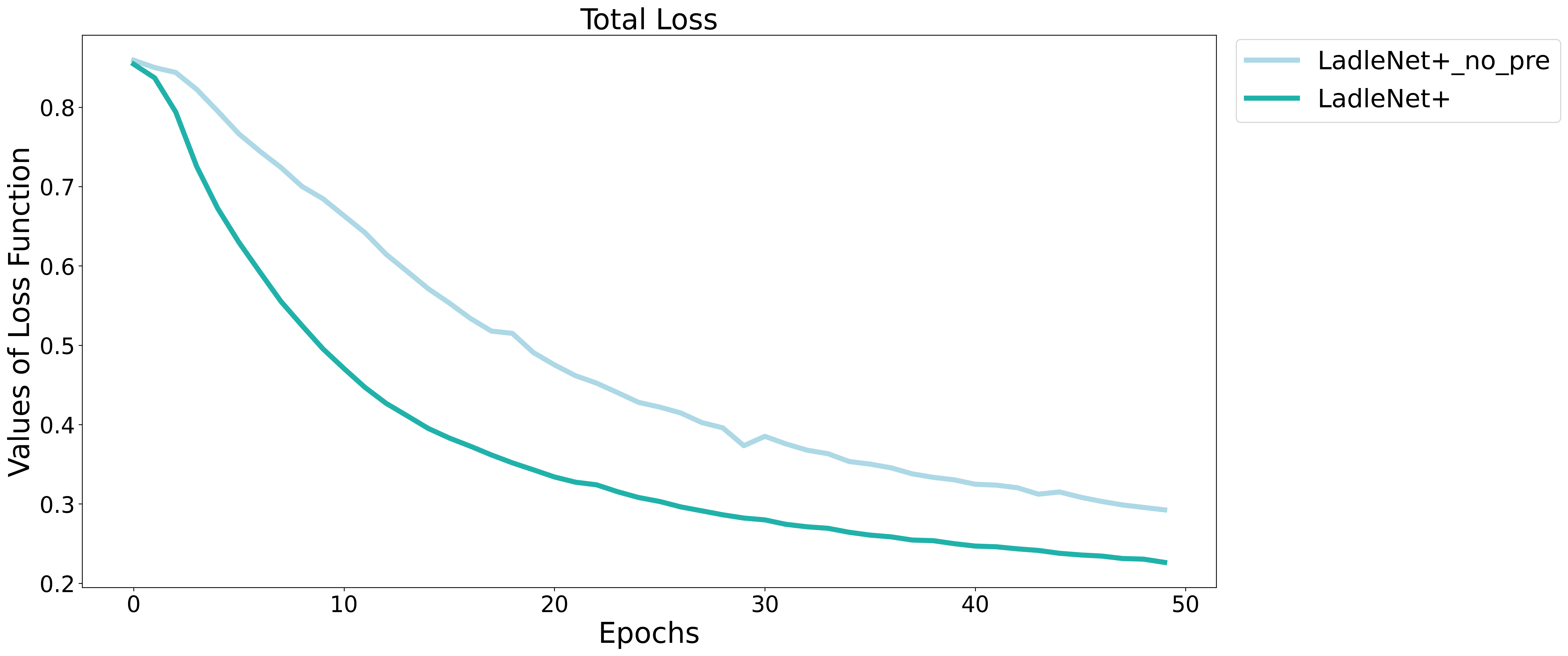}
    \caption{Comparative results of pre-training methods for the speed of convergence of model loss values.}
    \label{fig:11}
\end{figure}

To validate the effectiveness of the semantic segmentation network in the Handle module, we conducted a set of ablation experiments to compare the impact of using a pretraining method on the performance of LadleNet+. Table \ref{tab:4} presents the results of the ablation experiments, while Figure \ref{fig:11} displays the convergence speed of the loss values for the two models. The results indicate that employing a pretrained semantic segmentation network can significantly enhance the model's performance and improve the convergence speed of the loss values. However, it's worth noting that LadleNet+ uses a pretrained model based on real VI images, rather than TIR images. We believe that utilizing a pretrained semantic segmentation model based on TIR images could lead to even greater improvements in model performance.

\section{Conclusion}
This paper introduces a novel network, LadleNet, which concatenates two U-net models, leveraging skip connections and feature aggregation to substantially enhance model performance. Additionally, we propose an extension of LadleNet, named LadleNet+, which improves the overall model's semantic space construction capability by replacing the backbone network of LadleNet. This augmentation aims to generate more realistic translation results. Both proposed approaches are evaluated on publicly available datasets, producing perceptually convincing images. In quantitative experiments, our models exhibit advanced performance in both macro-level indicators and image quality metrics. Furthermore, ablation experiments are conducted to validate the effectiveness of our proposed model improvements and extensions. The results from these experiments demonstrate a significant boost in model performance, underscoring the flexibility of our model's extensibility. We believe that in future research, the translation of TIR images to VI images will become a focal point, enhancing registration and fusion methods for TIR and VI images, thus yielding more robust model performance.

\printbibliography

\end{CJK}

\end{document}